\documentclass[sigconf,authorversion]{acmart} % 

\AtBeginDocument{%
  \providecommand\BibTeX{{%
    \normalfont B\kern-0.5em{\scshape i\kern-0.25em b}\kern-0.8em\TeX}}}

\copyrightyear{2022}
\acmYear{2022}
\setcopyright{acmcopyright}
\acmConference[KDD '22] {Proceedings of the 28th ACM SIGKDD Conference on Knowledge Discovery and Data Mining}{August 14--18, 2022}{Washington, DC, USA.}
\acmBooktitle{Proceedings of the 28th ACM SIGKDD Conference on Knowledge Discovery and Data Mining (KDD '22), August 14--18, 2022, Washington, DC, USA}
\acmPrice{15.00}
\acmISBN{978-1-4503-9385-0/22/08}
\acmDOI{10.1145/3534678.3539296}
\graphicspath{{figures/}}
\usepackage{tikz}
\def\checkmark{\tikz\fill[scale=0.4](0,.35) -- (.25,0) -- (1,.7) -- (.25,.15) -- cycle;}

\begin{document}

\newcommand{\textbs}[1]{\textcolor{red}{\textbf{#1}}}
\newcommand{\textgd}[1]{\textcolor{violet}{#1}}
\newcommand{\bigparallel}{\bigm\Vert}

% \title{Edge-Augmented Graph Transformers}
\title{Global Self-Attention as a Replacement for Graph Convolution}

\author{Md Shamim Hussain}
\email{hussam4@rpi.edu}
\orcid{0000-0002-0832-913X}
\affiliation{%
  \institution{Rensselaer Polytechnic Institute}
  \city{Troy}
  \state{New York}
  \country{USA}
}

\author{Mohammed J. Zaki}
\email{zaki@cs.rpi.edu}
\orcid{0000-0003-4711-0234}
\affiliation{%
  \institution{Rensselaer Polytechnic Institute}
  \city{Troy}
  \state{New York}
  \country{USA}
}

\author{Dharmashankar Subramanian}
\email{dharmash@us.ibm.com}
\orcid{0000-0002-1990-7740}
\affiliation{%
  \institution{IBM T. J. Watson Research Center}
  \city{Yorktown Heights}
  \state{New York}
  \country{USA}}

\renewcommand{\shortauthors}{Hussain, Zaki and Subramanian}

\begin{abstract}
  We propose an extension to the transformer neural network architecture for general-purpose graph learning by adding a dedicated pathway for pairwise structural information, called edge channels. The resultant framework -- which we call Edge-augmented Graph Transformer (EGT) -- can directly accept, process and output structural information of arbitrary form, which is important for effective learning on graph-structured data. Our model exclusively uses global self-attention as an aggregation mechanism rather than static localized convolutional aggregation. This allows for unconstrained long-range dynamic interactions between nodes. Moreover, the edge channels allow the structural information to evolve from layer to layer, and prediction tasks on edges/links can be performed directly from the output embeddings of these channels. We verify the performance of EGT in a wide range of graph-learning experiments on benchmark datasets, in which it outperforms Convolutional/Message-Passing Graph Neural Networks. EGT sets a new state-of-the-art for the quantum-chemical regression task on the OGB-LSC PCQM4Mv2 dataset containing 3.8 million molecular graphs. Our findings indicate that global self-attention based aggregation can serve as a flexible, adaptive and effective replacement of graph convolution for general-purpose graph learning. Therefore, convolutional local neighborhood aggregation is not an essential inductive bias.
\end{abstract}

\begin{CCSXML}
  <ccs2012>
     <concept>
      <concept_id>10010147.10010257.10010293.10010294</concept_id>
      <concept_desc>Computing methodologies~Neural networks</concept_desc>
      <concept_significance>500</concept_significance>
      </concept>
     <concept>
         <concept_id>10010147.10010178</concept_id>
         <concept_desc>Computing methodologies~Artificial intelligence</concept_desc>
         <concept_significance>300</concept_significance>
         </concept>
   </ccs2012>
\end{CCSXML}
  
\ccsdesc[500]{Computing methodologies~Neural networks}
\ccsdesc[300]{Computing methodologies~Artificial intelligence}

\keywords{graph neural networks, graph representation learning, self-attention}


\maketitle

\section{Introduction}
Graph-structured data are ubiquitous in different areas such as communication networks, molecular structures, citation networks, knowledge bases and social networks. Due to the flexibility of the structural information in graphs, they are powerful tools for compact and intuitive representation of data originating from a very wide range of sources. However, this flexibility comes at the cost of added complexity in processing and learning from graph-structured data, due to the arbitrary nature of the interconnectivity of the nodes. Recently the go-to solution for deep representation learning on graphs has been Graph Neural Networks (GNNs) \citep{gori2005new,scarselli2008graph}. The most commonly used GNNs follow a convolutional pattern whereby each node in the graph updates its state based on that of its neighbors \cite{kipf2016semi,xu2018powerful} in each layer. On the other hand, the pure self-attention based transformer architecture \citep{vaswani2017attention} has displaced convolutional neural networks for more regularly arranged data, such as sequential (e.g., text) and grid-like (images) data, to become the new state-of-the-art, especially in large-scale learning. Transformers have become the de-facto standard in the field of natural language processing, where they have achieved great success in a wide range of tasks such as language understanding, machine translation and question answering. The success of transformers has translated to other forms of unstructured data in different domains such as audio \citep{child2019generating, li2019neural} and images \citep{chen2020generative, dosovitskiy2020image} and also on different (classification/generation, supervised/unsupervised) tasks.

Transformers differ from convolutional neural networks in some important ways. A convolutional layer aggregates a localized window around each position to produce an output for that position. The weights that are applied to the window are independent of the input, and can therefore be termed as \emph{static}. Also, the sliding/moving window directly follows the structure of the input data, i.e., the sequential or grid-like pattern of positions. This is an apriori assumption based on the nature of the data and how it should be processed, directly inspired by the filtering process in signal processing. We call this assumption the \emph{convolutional inductive bias}. On the other hand, in the case of a transformer encoder layer, the internal arrangement of the data does not directly dictate how it is processed. Attention weights are formed based on the queries and the keys formed at each position, which in turn dictate how each position aggregates other positions. The aggregation pattern is thus global and input dependent, i.e., it is \emph{dynamic}. The positional information is treated as an input to the network in the form of positional encodings. In their absence, the transformer encoder is permutation equivariant and treats the input as a multiset. Information is propagated among different positions only via the global self-attention mechanism, which is agnostic to the internal arrangement of the data. Due to this property of global self-attention, distant points in the data can interact with each other as efficiently as nearby points. Also, the network \emph{learns} to form appropriate aggregation patterns during the training process, rather than being constrained to a predetermined pattern.

Although it is often straightforward to adopt the transformer architecture for regularly structured data such as text and images by employing an appropriate positional encoding scheme, the highly arbitrary nature of structure in graphs makes it difficult to represent the position of each node only in terms of positional encodings. Also, it is not clear how edge features can be incorporated in terms of node embeddings. For graph-structured data, the edge/structural information can be just as important as the node information, and thus we should expect the network to process this information hierarchically, just like the node embeddings. To facilitate this, we introduce a new addition to the transformer, namely \emph{residual edge channels} -- a pathway that can leverage structural information. This is a simple yet powerful extension to the transformer framework in that it allows the network to directly process graph-structured data. This addition is also very general in the sense that it facilitates the input of structural information of arbitrary form, including edge features, and can handle different variants of graphs such as directed and weighted graphs in a systematic manner. Our framework can exceed the results of widely used Graph Convolutional Networks on datasets of moderate to large sizes, in supervised benchmarking tasks while maintaining a similar number of parameters. But our architecture deviates significantly from convolutional networks in that it does not impose any strong inductive bias such as the convolutional bias, on the feature aggregation process. We rely solely on the global self-attention mechanism to learn how best to use the structural information, rather than constraining it to a fixed pattern. Additionally, the structural information can evolve over layers and the network can potentially form new structures. Any prediction on the structure of the graph, such as link prediction or edge classification, can be done directly from the outputs of edge channels. However, these channels do add to the quadratic computational and memory complexity of global self-attention, with respect to the number of nodes, which restricts us to moderately large graphs. In addition to the edge channels, we generalize GNN concepts like gated aggregation \citep{bresson2017residual}, degree scalers \citep{corso2020principal} and positional encodings \cite{dwivedi2020benchmarking} for our framework.

\begin{figure}[!t]
  \centering
  \includegraphics[width=.934\columnwidth]{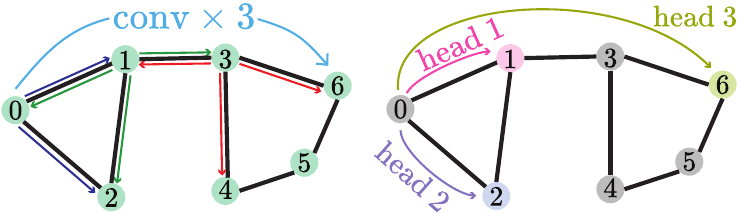}
  \caption{A conceptual demonstration of Graph Convolution (left) and Global Self-Attention (right). It takes three stages of convolution for node 0 to aggregate node 6. With global self-attention, the model can learn to do so in a single step. The attention heads are formed dynamically for a given graph.}
  \Description{A diagram showing conceptually how global self-attention differs from graph convolution, emphasizing on its far-sightedness and dynamic nature.}
  \label{fig:conv_v_sa}
\end{figure}

Our experimental results indicate that given enough data and with the proposed edge channels, the model can utilize global self-attention to learn the best aggregation pattern for the task at hand. Thus, our results indicate that following a fixed convolutional aggregation pattern whereby each node is limited to aggregating its closest neighbors (based on adjacency, distance, intimacy, etc.) is not an essential inductive bias. With the flexibility of global self-attention, the network can learn to aggregate distant parts of the input graph in just one step as illustrated in Fig.~\ref{fig:conv_v_sa}. Since this pattern is learned rather than being imposed by design, it increases the expressivity of the model. Also, this aggregation pattern is dynamic and can adapt to each specific input graph. Similar findings have been reported for unstructured data such as images \citep{dosovitskiy2020image, cordonnier2019relationship, ramachandran2019stand}. Some recent works have reported global self-attention as a means for better generalization or performance by improving the expressivity of graph convolutions \citep{puny2020global,wang2021global}. Very recently, Graphormer \citep{ying2021transformers} performed well on graph level prediction tasks on molecular graphs by incorporating edges with specialized encodings. However, it does not directly process the edge information and therefore does not generalize well to edge-related prediction tasks. By incorporating the edge-channels, we are the first to propose global self-attention as a direct and general replacement for graph convolution for node-level, link(edge)-level and graph-level prediction, on all types of graphs.

\section{Related Work}
In relation to our work, we discuss self-attention based GNN models, where the attention mechanism is either constrained to a local neighborhood (local self-attention) of each node or unconstrained over the whole input graph (global self-attention). Methods like Graph Attention Network (GAT) \citep{velivckovic2017graph} and Graph Transformer (GT) \citep{dwivedi2020generalization} constrain the self-attention mechanism to local neighborhoods of each node only, which is reminiscent of the graph convolution/local message-passing process. Several works have attempted to adopt the global self-attention mechanism for graphs as well. Graph-BERT \citep{zhang2020graph} uses a modified transformer framework on a sampled linkless subgraph (i.e., only node representations are processed) around a target node. Since the nodes do not inherently bear information about their interconnectivity, Graph-BERT uses several types of relative positional encodings to embed the information about the edges within a subgraph. Graph-BERT focuses on unsupervised representation learning by training the model to predict a single masked node in a sampled subgraph. GROVER \citep{rong2020self} used a modified transformer architecture with queries, keys and values produced by Message-Passing Networks, which indirectly incorporate the input structural information. This framework was used to perform unsupervised learning on molecular graphs only. Graph Transformer \citep{cai2020graph} and Graphormer \cite{ying2021transformers} directly adopt the transformer framework for specific tasks. Graph Transformer separately encodes the nodes and the relations between nodes to form a fully connected view of the graph which is incorporated into a transformer encoder-decoder framework for graph-to-sequence learning. Graphormer incorporates the existing structure/edges in the graph as an attention bias, formed from the shortest paths between pairs of nodes. It focuses on graph-level prediction tasks on molecular graphs (e.g., classification/regression on molecular graphs). Unlike these models which handle graph structure in an ad-hoc manner and only for a specific problem, we directly incorporate graph structure into the transformer model via the edge channels and propose a general-purpose learning framework for graphs based only on the global self-attention mechanism, free of the strong inductive bias of convolution. Apart from being used for node feature aggregation, attention has also been used to form metapaths in heterogeneous graphs, such as the Heterogeneous Graph Transformer (HGT) \citep{hu2020heterogeneous} and the Graph Transformer network (GTN) \citep{yun2019graph}. However, these works are orthogonal to ours since metapaths are only relevant in the case of heterogeneous graphs and these methods use attention \emph{specifically} to combine heterogeneous edges, over multiple hops. We focus only on homogeneous graphs, but more importantly, we use attention as a global aggregation mechanism. 

\section{Network Architecture}
\subsection{Preliminaries}
The transformer architecture was proposed by \citet{vaswani2017attention} as a purely attention-based model. The transformer encoder uses self-attention to communicate information between different positions, and thus produces the output embeddings for each position. In the absence of positional encodings, this process is permutation equivariant and treats the input embeddings as a multiset.

Each layer in the transformer encoder consists of two sublayers. The key component of the transformer is the multihead self-attention mechanism which takes place in the first sublayer, which can be expressed as:
\begin{align}
  \mathrm{Attn}(\mathbf{Q}, \mathbf{K}, \mathbf{V}) &=\tilde{\mathbf{A}}\mathbf{V} \label{eq:xformer_enc1} \\
  \text{Where,} \ \ \ \tilde{\mathbf{A}} &= \mathrm{softmax}\left(\frac{\mathbf{Q} \mathbf{K}^T}{\sqrt{d_k}}\right) \label{eq:xformer_enc2} 
\end{align}

where $\mathbf{Q}, \mathbf{K}, \mathbf{V}$ are the keys, queries and values formed by learned linear transformations of the embeddings and $d_k$ is the dimensionality of the queries and the keys. $\tilde{\mathbf{A}}$ is known as the (softmax) attention matrix, formed from the scaled dot product of queries and keys. This process is done for multiple sets of queries, keys and values, hence the name multihead self-attention. The second sublayer is the feedforward layer which serves as a pointwise non-linear transformation of the embeddings.

\subsection{Edge-augmented Graph Transformer (EGT)}
\begin{figure}[!t]
  \centering
  \includegraphics[height=0.464\textheight]{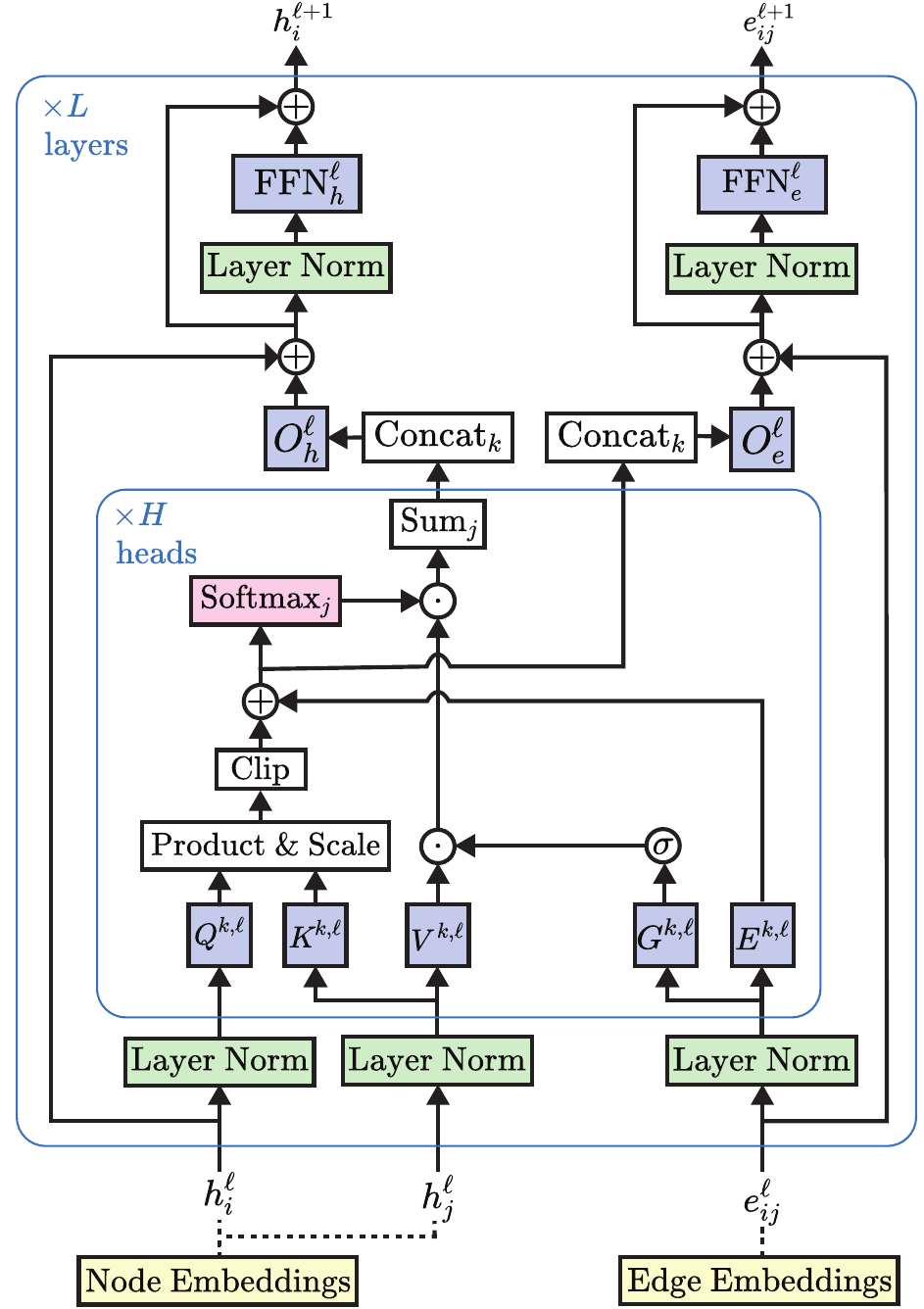}
  \caption{Edge-augmented Graph Transformer (EGT)}
  \Description{The network architection of Edge-augmented Graph Transformer (EGT) which is an extension of the transformer architecture for general purpose graph learning.}
  \label{fig:arch}
\end{figure}
The EGT architecture (Fig.~\ref{fig:arch}) extends the original transformer architecture. The permutation equivariance of the transformer is ideal for processing the node embeddings in a graph because a graph is invariant under the permutation of the nodes, given that the edges are preserved. We call the residual channels present in the original transformer architecture \emph{node channels}. These channels transform a set of input node embeddings $\{h_1^{0}, h_2^{0}, ..., h_N^{0}\}$ into a set of output node embeddings $(h_i^{L})_{final}$ (for $1\le i \le N$), where  $h_i^{\ell} \in \mathbb{R}^{d_h}$, $d_h$ is the node embeddings dimensionality,  $N$ is the number of nodes, and $L$ is the number of layers. Our contribution to the transformer architecture is the introduction of \emph{edge channels}, which start with an embedding for each \emph{pair of nodes}. Thus, there are $N \times N$ input edge embeddings $e_{11}^0, e_{12}^0, ..., e_{1N}^0, e_{21}^0, ..., e_{NN}^0$ where, $e_{ij}^l \in \mathbb{R}^{d_e}$, $d_e$ is the edge embeddings dimensionality. The input edge embeddings are formed from graph structural matrices and edge features. We define a graph structural matrix as any matrix with dimensionality $N\times N$, which can completely or partially define the structure of a graph (e.g., adjacency matrix, distance matrix). The edge embeddings are updated by EGT in each layer and finally, it produces a set of output edge embeddings $(e_{ij}^L)_{final}$ (for $1\le i,j \le N$) from which structural predictions such as edge labeling and link prediction can be performed.

From equations (\ref{eq:xformer_enc1}) and (\ref{eq:xformer_enc2}) we see that the attention matrix is comparable with a row-normalized adjacency matrix of a directed weighted complete graph. It dictates how the node features in a graph are aggregated, similarly to GCN \citep{kipf2016semi}. Unlike the input graph, this graph is dynamically formed by the attention mechanism. However, the basic transformer does not have a direct way to incorporate the input structure (existing edges) while forming these weighted graphs, i.e., the attention matrices. Also, these dynamic graphs are collapsed immediately after the aggregation process is done. To remedy the first problem we let the edge channels participate in the aggregation process as follows (as shown in Fig.~\ref{fig:arch}) -- in the $\ell$'th layer and for the $k$'th attention head,
\begin{align}
  \mathrm{Attn}(\mathbf{Q}_h^{k,\ell}, \mathbf{K}_h^{k,\ell}, \mathbf{V}_h^{k,\ell}) &= \tilde{\mathbf{A}}^{k,\ell} \mathbf{V}_h^{k,\ell} \label{eq:mod_attn}\\
  \text{Where,} \ \ \ \tilde{\mathbf{A}}^{k,\ell} &= \mathrm{softmax} ( \hat{\mathbf{H}}^{k,\ell} ) \odot \sigma(\mathbf{G}^{k,\ell}_{e})  \\
  \text{Where,} \ \ \ \hat{\mathbf{H}}^{k,\ell}  &= \mathrm{clip}\left(\frac{\mathbf{Q}_h^{k,\ell} (\mathbf{K}_h^{k,\ell})^T}{\sqrt{d_k}}\right) + \mathbf{E}^{k,\ell}_{e}
\end{align}
where $\odot$ denotes elementwise product. $\mathbf{E}^{k,\ell}_{e}, \mathbf{G}^{k,\ell}_{e} \in \mathbb{R}^{N \times N}$ are concatenations of the learned linear transformed edge embeddings. $\mathbf{E}^{k,\ell}_{e}$ is a bias term added to the scaled dot product between the queries and the keys. It lets the edge channels influence the attention process. $\mathbf{G}_e^{k,\ell}$ drives the sigmoid $\sigma (\cdot)$ function and lets the edge channels also \emph{gate} the values before aggregation, thus controlling the flow of information between nodes. The scaled dot product is clipped to a limited range which leads to better numerical stability (we used $[-5,+5]$). To ensure that the network takes advantage of the full-connectivity the attention process is randomly masked by adding $-\infty$ to the inputs to the softmax with a small probability during training (i.e., random attention masking). Another approach is to apply dropout \cite{srivastava2014dropout} to the attention matrix.

To let the structural information evolve from layer to layer, the edge embeddings are updated by a learnable linear transformation of the inputs to the softmax function. The outputs of the attention heads are also mixed by a linear transformation. To facilitate training deep networks, Layer Normalization (LN) \citep{ba2016layer} and residual connections \citep{he2016deep} are used. We adopted the Pre-Norm architecture whereby normalization is done immediately before the weighted sublayers \citep{xiong2020layer} rather than after, because of its better optimization characteristics. So, $\hat{h}_i^{\ell} =  \mathrm{LN}(h_i^{\ell-1})$, $\hat{e}_{ij}^{\ell} = \mathrm{LN}(e_{ij}^{\ell-1})$. The residual updates can be expressed in an elementwise manner as:
\begin{align}
  \hat{\hat{h}}_i^{\ell} &= h_i^{\ell-1} + \mathbf{O}_h^\ell \bigparallel_{k=1}^{H} \sum_{j=1}^{N} \tilde{\mathbf{A}}_{ij}^{k,\ell} (\mathbf{V}^{k,\ell} \hat{h}_i^{\ell}) \label{eq:agg}\\
  \hat{\hat{e}}_{ij}^{\ell} &= e_{ij}^{\ell-1} + \mathbf{O}_e^\ell \bigparallel_{k=1}^{H} \hat{\mathbf{H}}_{ij}^{k,\ell}
\end{align}
Here, $\parallel$ denotes concatenation. $\mathbf{O}_h^\ell \in \mathbb{R}^{d_h \times d_h}$ and $\mathbf{O}_e^\ell \in \mathbb{R}^{d_e \times H}$ are the learned output projection matrices, with edge embeddings dimensionality $d_e$ and $H$ attention heads.

The feed-forward sublayer following the attention sublayer consists of two consecutive pointwise fully connected linear layers with a non-linearity such as ELU \citep{clevert2015fast} in between. The updated embeddings are $h_i^{\ell}=  \hat{\hat{h}}_i^{\ell} + \mathrm{FFN_h^\ell}(\mathrm{LN}(\hat{\hat{h}}_i^{\ell}))$, $e_i^{\ell}=  \hat{\hat{e}}_i^{\ell} + \mathrm{FFN_e^\ell}(\mathrm{LN}(\hat{\hat{e}}_i^{\ell}))$. The Pre-Norm architecture also ends with a layer normalization over the final embeddings as $(h_i^{L})_{final} = \mathrm{LN}(h_i^{L})$, $(e_{ij}^{L})_{final} = \mathrm{LN}(e_{ij}^{L})$.

\subsection{Dynamic Centrality Scalers}
The attention mechanism in equation (\ref{eq:mod_attn}) is a weighted average of the gated node values, which is agnostic to the degree of the nodes. However, we may want to make the network sensitive to the degree/centrality of the nodes, in order to make it more expressive when distinguishing between non-isomorphic (sub-)graphs, similar to GIN \cite{xu2018powerful}. While this can be achieved by directly encoding the degrees of the nodes as an additional input like \citep{ying2021transformers}, we aimed for an approach that is adaptive to the dynamic nature of self-attention. \citet{corso2020principal} propose scaling the aggregated values by a function of the degree of the node, more specifically a logarithmic degree scaler. But it is tricky to form a notion of degree/centrality for the dynamically formed graph represented by the attention matrix because this row-normalized matrix bears no notion of degree. In our network, the sigmoid gates control the flow of information to a particular node which are derived from the edge embeddings. So we use the sum of the sigmoid gates as a measure of centrality for a node and scale the aggregated values by the logarithm of this sum. With centrality scalers, equation (\ref{eq:agg}) becomes:
\begin{align}
  \hat{\hat{h}}_i^{\ell} &= h_i^{\ell-1} + \mathbf{O}_h^\ell \bigparallel_{k=1}^{H} s^{k,l}_i \sum_{j=1}^{N} \tilde{\mathbf{A}}_{ij}^{k,\ell} (\mathbf{V}^{k,\ell} \hat{h}_i^{\ell}) \\
  \text{Where,} \ \ \ s^{k,l}_i &= \ln \left(1 + \sum_{j=1}^{N} \sigma(\mathbf{G}^{k,\ell}\mathbf{e}_{i,j}) \right)
\end{align}
Here, $s^{k,l}_i$ is the centrality scaler for node $i$, for attention head $k$ at layer $\ell$. As pointed out by \citet{ying2021transformers}, with the addition of a centrality measure the global self-attention mechanism becomes at least as powerful as the 1-Weisfeiler-Lehman (1-WL) isomorphism test and potentially even more so, due to aggregation over multiple hops. Note that commonly used convolutional GNNs like GIN are at most as powerful as the 1-WL isomorphism test \cite{xu2018powerful}.

\subsection{SVD-based Positional Encodings}
While applying the transformer on regularly arranged data such as sequential (e.g., text) and grid-like (e.g., images) data it is customary to use sinusoidal positional encodings introduced by \citet{vaswani2017attention}. However, the arbitrary nature of structure in graphs makes it difficult to devise a consistent positional encoding scheme. Nonetheless, positional encodings have been used for GNNs to embed global positional information within individual nodes and to distinguish isomorphic nodes and edges \citep{murphy2019relational,srinivasan2019equivalence}. Inspired by matrix factorization based node embedding methods for graphs \citep{belkin2001laplacian}, \citet{dwivedi2020benchmarking} proposed to use the $k$ smallest non-trivial eigenvectors of the Laplacian matrix of the graph as positional encodings. However, since the Laplacian eigenvectors can be complex-valued for directed graphs, this method is more relevant for undirected graphs which have symmetric Laplacian matrices. To remedy this we propose a method, that is more general and applies to all variants of graphs (e.g., directed, weighted). We propose a form of positional encoding based on precalculated SVD of the graph structural matrices. We use the largest $r$ singular values and corresponding left and right singular vectors to form our positional encodings. We use the adjacency matrix $\mathbf{A}$ (with self-loops) as the graph structural matrix, but it can be generalized to other structural matrices since the SVD of any real matrix produces real singular values and vectors.
\begin{align}
  \mathbf{A} &\stackrel{\mathrm{SVD}}{\approx}
  \mathbf{U}\mathbf{\Sigma}\mathbf{V}^T 
  = (\mathbf{U}\sqrt{\Sigma}) \cdot (\mathbf{V}\sqrt{\Sigma})^T
  = \hat{\mathbf{U}}\hat{\mathbf{V}}^T \label{eq:svd}\\
  \hat{\mathbf{\Gamma}} &= \hat{\mathbf{U}} \parallel \hat{\mathbf{V}}
\end{align}
Where $\mathbf{U,V}\in \mathbb{R}^{N\times r}$ matrices contain the $r$ left and right singular vectors as columns, respectively, corresponding to the top $r$ singular values in the diagonal matrix $\mathbf{\Sigma} \in \mathbb{R}^{r \times r}$. Here, $\parallel$ denotes concatenation along the columns. From (\ref{eq:svd}) we see that the dot product between $i$'th row of $\hat{\mathbf{U}}$ and $j$'th row of $\hat{\mathbf{V}}$ can approximate $\mathbf{A}_{ij}$ which denotes whether there is an edge between nodes $i$ and $j$. Thus, the rows of $\hat{\mathbf{\Gamma}}$, namely $\hat{\mathbf{\gamma}}_1, \hat{\mathbf{\gamma}}_2, ..., \hat{\mathbf{\gamma}}_N$, each with dimensionality $\hat{\mathbf{\gamma}}_i \in \mathbb{R}^{2r}$, bear denoised information about the edges and can be used as positional encodings. Note that this form of representation based on the dot product is consistent with the scaled dot product attention used in the transformer framework. Since the signs of corresponding pairs of left and right singular vectors can be arbitrarily flipped, we randomly flip the signs of $\hat{\mathbf{\gamma}}_i$ during training for better generalization. Instead of directly adding $\hat{\mathbf{\gamma}}_i$ to the input embeddings of the node $i$, we add a projection $\mathbf{\gamma}_i=\mathbf{W}_{enc}\hat{\mathbf{\gamma}}_i$, where $\mathbf{W}_{enc}\in \mathbb{R}^{d_h \times 2r}$ is a learned projection matrix. This heuristically leads to better results. Since our architecture directly takes structure as input via the edge channels, the inclusion of positional encodings is optional for most tasks. However, positional encodings can help distinguish isomorphic nodes \citep{zhang2020revisiting} by serving as an absolute global coordinate system. Thus, they make the model more expressive. However, the absolute coordinates may, in theory, hamper generalization, because they are specific to a particular reference frame that depends on the input graph. But in practice, we did not find any detrimental effect on the performance for any task.

\subsection{Embedding and Prediction}
Given an input graph, both node and edge feature embeddings are formed by performing learnable linear transformations for continuous vector values, or vector embeddings for categorical/discrete values. In the case of multiple sets of features, their corresponding embeddings are added together. When positional encodings $\mathbf{\gamma}_i$ are used, they are added to the input node embeddings. The edge embeddings are formed by adding together the embeddings from the graph structural matrix and the input edge feature embeddings (when present). For non-existing edges, a masking value/vector is used in the place of an edge feature. As input structural matrix, we use the distance matrix clipped up to $k$-hop distance, i.e., $\mathbf{D}^{(k)}$ where $\mathbf{D}^{(k)}_{ij} \in \{0,1,...,k\}$ are the shortest distances between nodes $i$ and $j$, clipped to a maximum value of $k$. We use vector embeddings of the discrete values contained in these matrices.

For node and edge classification/regression tasks, we apply a few final MLP layers on the final node and edge embeddings, respectively, to produce the output. For graph-level classification/reg-ression we adopt one of two different methods. In \emph{global average pooling} method, all the output node embeddings are averaged to form a graph-level embedding, on which final linear layers are applied. In \emph{virtual nodes} method, $q$ new virtual nodes with learnable input embeddings $h_{N+1}^{0}, h_{N+2}^{0}, ..., h_{N+q}^{0}$ are passed through EGT along with existing node embeddings. There are also $q$ different learnable edge embeddings $\tilde{e}_i$ which are used as follows -- the edge embedding between a virtual node $i$ and existing graph node $j$ is assigned $e_{ij}^0=e_{ji}^0=\tilde{e}_i$, and the edge embeddings between two virtual nodes $i,j$, are assigned $e_{ij}^0=e_{ji}^0=\frac{1}{2}(\tilde{e}_i+\tilde{e}_j)$. Finally, the graph embedding is formed by concatenating the output node embeddings of the virtual nodes. This method is more flexible and better suited for larger models. The centrality scalers mentioned above are not applied to the virtual nodes, because by nature these nodes have high levels of centrality which are very different from the graph nodes. So a fixed scaler value of $s^{k,l}_i=1$ is used instead for these virtual nodes.

For smaller datasets, we found that adding a secondary \emph{distance prediction objective} alongside the graph-level prediction task in a multi-task learning setting serves as an effective means of regularization and thus improves the generalization of the trained model. This self-supervised objective is reminiscent of the unsupervised link prediction objective often used to pre-train GNNs to form node embeddings. In our case, we take advantage of the fact that we have output edge embeddings from the edge channels (alongside the node embeddings, which are used for graph-level prediction). We thus pass the output edge embeddings through a few (we used three) MLP layers and set the distance matrix up to $\nu$-hop, $\mathbf{D}^{(\nu)}$, as a categorical target. Hops greater than $\nu$ are ignored while calculating the loss. The loss from this secondary objective is multiplied by a small factor $\kappa$ and added to the total loss. Note that in this case we always use the adjacency matrix, rather than the distance matrix as the input graph structural matrix so that the edge channels do not simply learn an identity transformation. We emphasize that this objective is only potentially beneficial as a regularization method for smaller datasets by guiding the aggregation process towards a Breadth-First Search pattern, which is a \emph{soft} form of the convolutional bias. In the presence of enough data, the network is able to learn the best aggregation pattern for the given primary objective, which also generalizes to unseen data.

\section{Experiments and Results}
\newcommand{\citeGCN}{\citep{kipf2016semi}}
\newcommand{\citeGraphSage}{\citep{hamilton2017representation}}
\newcommand{\citeGIN}{\citep{xu2018powerful}}
\newcommand{\citeGAT}{\citep{velivckovic2017graph}}
\newcommand{\citeGT}{\citep{dwivedi2020generalization}}
\newcommand{\citeGatedGCN}{\citep{bresson2017residual}}
\newcommand{\citePNA}{\citep{corso2020principal}}
\newcommand{\citeDGN}{\citep{beani2021directional}}
\newcommand{\citeGraphormer}{\citep{ying2021transformers}}
\begin{table*}[!ht]
  \centering
  \caption{Experimental results on 6 benchmarking datasets from \citet{dwivedi2020benchmarking}. Results on PATTERN and CLUSTER datasets are given in terms of weighted accuracy. \textbs{Red:} best model, \textgd{Violet:} good model; arrow next to a metric indicates whether higher or lower is better. Results not shown are not available for that method.}
  \scalebox{0.77}{
    \begin{tabular}{l|cc|c|c|c|cc|cc}
    \toprule
                        %  & \multicolumn{3}{c|}{\textbf{Node Classification}}                                             & \multicolumn{2}{c|}{\textbf{Graph Classification}}            & \multicolumn{2}{c|}{\textbf{Edge Classification}}             & \multicolumn{2}{c}{\textbf{Graph Regression}}                 \\
    % \cmidrule{2-10}
                                         & \multicolumn{2}{c|}{\textbf{PATTERN}}                         & \textbf{CLUSTER}                & \textbf{MNIST}                  & \textbf{CIFAR10}                & \multicolumn{2}{c|}{\textbf{TSP}}                             & \multicolumn{2}{c}{\textbf{ZINC}}                             \\
                                         & \multicolumn{2}{c|}{\textbf{\% Accuracy} $\uparrow$}          & \textbf{\% Accuracy} $\uparrow$ & \textbf{\% Accuracy} $\uparrow$ & \textbf{\% Accuracy} $\uparrow$ & \multicolumn{2}{c|}{\textbf{F1} $\uparrow$}                   & \multicolumn{2}{c}{\textbf{MAE} $\downarrow$}                 \\
    \cmidrule{2-10}
                                         & \textbf{\#Param}              & \textbf{\#Param}              & \textbf{\#Param}                & \textbf{\#Param}                & \textbf{\#Param}                & \textbf{\#Param}              & \textbf{\#Param}              & \textbf{\#Param}              & \textbf{\#Param}              \\
    \textbf{Model}                       & \textbf{$\approx$100K}        & \textbf{$\approx$500K}        & \textbf{$\approx$500K}          & \textbf{$\approx$100K}          & \textbf{$\approx$100K}          & \textbf{$\approx$100K}        & \textbf{$\approx$500K}        & \textbf{$\approx$100K}        & \textbf{$\approx$500K}        \\
    \midrule\midrule                                                                     
    GCN \citeGCN                         &         63.880 $\pm$ 0.074    &         71.892 $\pm$ 0.334    &         68.498 $\pm$ 0.976      &         90.705 $\pm$ 0.218      &         55.710 $\pm$ 0.381      &         0.630  $\pm$ 0.001    &                               &          0.459 $\pm$ 0.006    &          0.367 $\pm$ 0.011    \\
    GraphSage \citeGraphSage             &         50.516 $\pm$ 0.001    &         50.492 $\pm$ 0.001    &         63.844 $\pm$ 0.110      &         97.312 $\pm$ 0.097      &         65.767 $\pm$ 0.308      &         0.665  $\pm$ 0.003    &                               &          0.468 $\pm$ 0.003    &          0.398 $\pm$ 0.002    \\
    GIN \citeGIN                         &         85.590 $\pm$ 0.011    &         85.387 $\pm$ 0.136    &         64.716 $\pm$ 1.553      &         96.485 $\pm$ 0.097      &         55.255 $\pm$ 1.527      &         0.656  $\pm$ 0.003    &                               &          0.387 $\pm$ 0.015    &          0.526 $\pm$ 0.051    \\
    GAT \citeGAT                         &         75.824 $\pm$ 1.823    &         78.271 $\pm$ 0.186    &         70.587 $\pm$ 0.447      &         95.535 $\pm$ 0.205      &         64.223 $\pm$ 0.455      &         0.671  $\pm$ 0.002    &                               &          0.475 $\pm$ 0.007    &          0.384 $\pm$ 0.007    \\
    GT \citeGT                           &                               &         84.808 $\pm$ 0.068    &         73.169 $\pm$ 0.622      &                                 &                                 &                               &                               &                               &          0.226 $\pm$ 0.014    \\
    GatedGCN \citeGatedGCN               &         84.480 $\pm$ 0.122    &         86.508 $\pm$ 0.085    & \textgd{76.082 $\pm$ 0.196}     &         97.340 $\pm$ 0.143      &         67.312 $\pm$ 0.311      & \textgd{0.808  $\pm$ 0.003}   & \textgd{0.838  $\pm$ 0.002}   &          0.375 $\pm$ 0.003    &          0.214 $\pm$ 0.013    \\
    PNA \citePNA                         &         86.567 $\pm$ 0.075    &                               &                                 &         97.690 $\pm$ 0.022      & \textgd{70.350 $\pm$ 0.630}     &                               &                               &          0.188 $\pm$ 0.004    &          0.142 $\pm$ 0.010    \\
    DGN \citeDGN                         & \textgd{86.680 $\pm$ 0.034}   &                               &                                 &                                 & \textbs{72.700 $\pm$ 0.540}     &                               &                               &  \textgd{0.168 $\pm$ 0.003}   &                               \\
    Graphormer \citeGraphormer           &                               & \textgd{86.650 $\pm$ 0.033}   &         74.660 $\pm$ 0.236      & \textgd{97.905 $\pm$ 0.176}     &         65.978 $\pm$ 0.579      &                               &         0.698  $\pm$ 0.007    &                               &  \textgd{0.122 $\pm$ 0.006}   \\
    \midrule                                                                                                                                                                                                                                                                                                             
    EGT                                  & \textbs{86.816 $\pm$ 0.027}   & \textbs{86.821 $\pm$ 0.020}   & \textbs{79.232 $\pm$ 0.348}     & \textbs{98.173 $\pm$ 0.087}     &         68.702 $\pm$ 0.409      & \textbs{0.822 $\pm$ 0.000}    &  \textbs{0.853 $\pm$ 0.001}   &  \textbs{0.143 $\pm$ 0.011}   &  \textbs{0.108 $\pm$ 0.009}   \\
    \bottomrule
\end{tabular}
  }
  \label{tab:results}%
\end{table*}%
We evaluate the performance of our proposed EGT architecture in a supervised and inductive setting. We focus on a diverse set of supervised learning tasks, namely, node and edge classification, and graph classification and regression. We also experiment on the transfer learning performance of EGT.

\noindent\textbf{Datasets:} In the medium-scale supervised learning setting, we experimented with the benchmarking datasets proposed by \citet{dwivedi2020benchmarking}, namely PATTERN (14K synthetic graphs, 44-188 \linebreak nodes/graph) and CLUSTER (12K synthetic graphs, 41-190 \linebreak nodes/graph) for node classification; TSP (12K synthetic graphs, 50-500 nodes/graph) for edge classification; and MNIST (70K superpixel graphs, 40-75 nodes/graph), CIFAR10 (60K superpixel graphs, 85-150 nodes/graph) and ZINC (12K molecular graphs, 9-37 \linebreak nodes/graph) for graph classification/regression. To evaluate the performance of EGT at large-scale we consider the graph regression task on the PCQM4M and its updated version PCQM4Mv2 datasets \cite{hu2021ogb} which contain 3.8 million molecular graphs with 1-51 nodes/graph. We also experimented on tranfer learning from PCQM4Mv2 dataset to the graph classification tasks on OGB \cite{hu2020open} datasets MolPCBA (438K molecular graphs, 1-332 nodes/graph) and MolHIV (41K molecular graphs, 2-222 nodes/graph).

\noindent\textbf{Evaluation Setup:} We use the PyTorch \cite{paszke2019pytorch} numerical library to implement our model. Training was done in a distributed manner on a single node with 8 NVIDIA Tesla V100 GPUs (32GB RAM/GPU), and 2 20-core 2.5GHz Intel Xeon CPUs (768GB RAM). Masked attention was used to process mini-batches containing graphs of different numbers of nodes. This allowed us to use highly parallel tensor operations on the GPU. The results are evaluated in terms of accuracy, F1 score, Mean Absolute Error (MAE), Average Precision (AP), or Area Under the ROC Curve (AUC), as recommended for each dataset. Hyperparameters were tuned on the validation set. Full details of hyperparameters are included in the appendix and the code is available at \url{https://github.com/shamim-hussain/egt}.

\subsection{Medium-scale Performance}
For the benchmarking datasets, we follow the training setting suggested by \citet{dwivedi2020benchmarking} and evaluate the performance of EGT for a given parameter budget. Comparative results are presented in Table~\ref{tab:results}. All datasets except PATTERN and CLUSTER include edge features. From the results, we see that EGT outperforms other GNNs (including GAT and GT which use local self-attention, and Graphormer which uses global self-attention but without edge channels) on all datasets except CIFAR10. We see a high level of overfitting for all models on CIFAR10, including our model which overfits the training dataset due to its higher capacity. The edge channels allow us to use the distance prediction objective in a multi-task learning setting, which helps lessen the overfitting problem on CIFAR10, ZINC and MNIST. Also, the output embeddings of the edge channels are directly used for edge classification on the TSP dataset which leads to very good results. Note that, Graphormer, which also uses global self-attention but does not have such edge channels, performs satisfactorily for other tasks but not so much on edge classification on the TSP dataset. Since we do not take advantage of the convolutional inductive bias our model shows various levels of overfitting on these medium-sized datasets. While EGT still outperforms other GNNs, we posit that it would further exceed the performance level of convolutional GNNs if more training data were present (we confirm this in the next section). Also, the results indicate that convolutional aggregation is not an essential inductive bias, and global attention can learn to make the best use of the structural information.

\newcommand{\citeGCNVN}{\citep{kipf2016semi,gilmer2017neural}}
\newcommand{\citeGINVN}{\citep{xu2018powerful,gilmer2017neural}}
\newcommand{\citeGINEVN}{\citep{brossard2020graph,gilmer2017neural}}
\newcommand{\citeDeeperGCNVN}{\citep{li2020deepergcn,gilmer2017neural}}
\subsection{Large-scale Performance}
\begin{table}[!t]
  \centering
  \caption{Results on OGB-LSC PCQM4M and PCQM4Mv2 datasets in terms of Mean Absolute Error (lower is better). Results not shown are not available.}
  \label{tab:pcqm4m_res}
  \scalebox{0.77}{
    \begin{tabular}{l|c|cc|cc}
    \toprule
                                                     &                     & \multicolumn{2}{c|}{\textbf{PCQM4M}}      & \multicolumn{2}{c}{\textbf{PCQM4Mv2}}             \\
    \cmidrule{3-6}               
    \textbf{Model}                                   & \textbf{\#Param}    & \textbf{Validate}         & \textbf{Test}     & \textbf{Validate}         & \textbf{Test-dev}         \\
    \midrule\midrule                                 
    GCN \citeGCN                                     & 2.0M                & 0.1684                    & 0.1838            & 0.1379                    & 0.1398                    \\
    GIN \citeGIN                                     & 3.8M                & 0.1536                    & 0.1678            & 0.1195                    & 0.1218                    \\
    GCN-VN \citeGCNVN                                & 4.9M                & 0.1510                    & 0.1579            & 0.1153                    & 0.1152                    \\
    GIN-VN \citeGINVN                                & 6.7M                & 0.1396                    & \textgd{0.1487}   & 0.1083                    & 0.1084                    \\
    GINE-VN \citeGINEVN                              & 13.2M               & 0.1430                    &                   &                           &                           \\
    DeeperGCN-VN \citeDeeperGCNVN                    & 25.5M               & 0.1398                    &                   &                           &                           \\
    GT \citeGT                                       & 0.6M                & 0.1400                    &                   &                           &                           \\
    GT (bigger model) \citeGT                        & 83.2M               & 0.1408                    &                   &                           &                           \\
    \midrule               
    Graphormer\textsubscript{SMALL} \citeGraphormer  & 12.5M               & 0.1264                    &                   &                           &                           \\
    Graphormer \citeGraphormer                       & 47.1M               & \textgd{0.1234}           & \textbs{0.1328}   & 0.0906                    &                           \\
    \midrule               
    EGT\textsubscript{Small} (6 layers)              & 11.5M               & 0.1260                    &                   & 0.0899                    &                           \\
    EGT\textsubscript{Medium} (18 layers)                                  & 47.4M               & \textbs{0.1224}           &                   & \textgd{0.0881}           &                           \\
    EGT\textsubscript{Large} (24 layers)             & 89.3M               &                           &                   & \textbs{0.0869}           & \textbs{0.0872}           \\
    % EGT\textsubscript{Larger} (30 layers)            & 110.8M              &                           &                   & \textbs{0.0869}           &                           \\
    \bottomrule
\end{tabular}
    }
\end{table}
\newcommand{\citeDeeperGCNVNFLAG}{\citep{li2020deepergcn,gilmer2017neural,kong2020flag}}
\newcommand{\citeDeeperGCNFLAG}{\citep{li2020deepergcn,kong2020flag}}
\newcommand{\citeGraphormerFLAG}{\citep{ying2020transformers,kong2020flag}}
\newcommand{\citePHC}{\citep{le2021parameterized}}
\begin{table}[!t]
  \centering
  \caption{Results on OGB Mol datasets. EGT uses transfer learning from PCQM4Mv2, whereas GIN-VN and Graphormer use transfer learning from PCQM4M. AP stands for Average Precision and AUC for Area Under the ROC Curve, higher is better for both. Results not shown are not available.}
  \label{tab:ogb_res}
  \scalebox{0.77}{
    \begin{tabular}{l|cc|cc}
    \toprule
                                                     & \multicolumn{2}{c|}{\textbf{MolPCBA}}                     & \multicolumn{2}{c}{\textbf{MolHIV}}                       \\
    \cmidrule{2-5}               
    \textbf{Model}                                   & \textbf{\#Param}    & \textbf{Test AP(\%)}                & \textbf{\#Param}    & \textbf{Test AUC(\%)}               \\
    \midrule\midrule                                 
    DeeperGCN-FLAG \citeDeeperGCNFLAG                & 6.55M               &         28.42 $\pm$ 0.43            & 532K                &         79.42 $\pm$ 1.20            \\
    DeeperGCN-VN-FLAG                                & 6.55M               &         28.42 $\pm$ 0.43            &                     &                                     \\
    \multicolumn{1}{r|}{\citeDeeperGCNVNFLAG}        &                     &                                     &                     &                                     \\
    PNA \citePNA                                     & 6.55M               &         28.38 $\pm$ 0.35            & 326K                &         79.05 $\pm$ 1.32            \\
    DGN \citeDGN                                     & 6.73M               &         28.85 $\pm$ 0.30            & 110K                & \textgd{79.70 $\pm$ 0.97}           \\
    GINE-VN \citeGINEVN                              & 6.15M               &         29.17 $\pm$ 0.15            &                     &                                     \\
    PHC-GNN \citePHC                                 & 1.69M               &         29.47 $\pm$ 0.26            & 114K                &         79.34 $\pm$ 1.16            \\
    \midrule                                                                                                    
    GIN-VN \citeGINVN                                & 3.4M                &         29.02 $\pm$ 0.17            & 3.3M                &         77.80 $\pm$ 1.82            \\
    (pre-trained)                                    &                     &                                     &                     &                                     \\
    \midrule
    Graphormer-FLAG \citeGraphormer                  & 119.5M              & \textbs{31.40 $\pm$ 0.34}           & 47.2M               & \textbs{80.51 $\pm$ 0.53}           \\
    (pre-trained)                                    &                     &                                     &                     &                                     \\
    \midrule               
    EGT\textsubscript{Larger} (30 layers)            & 110.8M              & \textgd{29.61 $\pm$ 0.24}           & 110.8M              & \textbs{80.60 $\pm$ 0.65}           \\
    (pre-trained)                                    &                     &                                     &                     &                                     \\
    \bottomrule
\end{tabular}
    }
\end{table}
\begin{table*}[!t]
  \centering
  \caption{Comparison of results for two ablated variants of EGT (EGT-Constrained and EGT-Simple), along with the complete architecture with (EGT) and without (EGT w/o PE) SVD based positional encodings}
  \scalebox{0.77}{
    \begin{tabular}{l|c|c|c|c|c|c|c}
    \toprule
                             &\textbf{PATTERN}                & \textbf{CLUSTER}                &\textbf{MNIST}                  & \textbf{CIFAR10}                & \textbf{TSP}                  & \textbf{ZINC}                 & \textbf{PCQM4Mv2}          \\
                             &\textbf{\% Accuracy} $\uparrow$ & \textbf{\% Accuracy} $\uparrow$ &\textbf{\% Accuracy} $\uparrow$ & \textbf{\% Accuracy} $\uparrow$ & \textbf{F1} $\uparrow$        & \textbf{MAE} $\downarrow$     & \textbf{MAE} $\downarrow$      \\
                             \cmidrule{2-8}
    \textbf{Model}           & \textbf{\#Param$\approx$500K}  & \textbf{\#Param$\approx$500K}   & \textbf{\#Param$\approx$100K}  & \textbf{\#Param$\approx$100K}   & \textbf{\#Param$\approx$500K} & \textbf{\#Param$\approx$500K} & \textbf{\#Param$\approx$11.5M} \\
    \midrule\midrule                                                                                                                                                                                                                                          
    EGT-Constrained          &         86.629 $\pm$ 0.041     &         76.701 $\pm$ 0.257      &          96.823 $\pm$ 0.204    & \textgd{65.192 $\pm$ 0.475}     & \textgd{0.846 $\pm$ 0.001}    & \textgd{0.174 $\pm$ 0.004}    &         0.0934                 \\
    EGT-Simple               & \textgd{86.813 $\pm$ 0.013}    & \textgd{79.182 $\pm$ 0.213}     &  \textgd{98.148 $\pm$ 0.139}   &         64.967 $\pm$ 1.263      &         0.831 $\pm$ 0.002     &         0.228 $\pm$ 0.020     & \textgd{0.0900}                \\                                                                  
    EGT w/o PE               & \textgd{86.812 $\pm$ 0.031}    &         77.665 $\pm$ 0.343      &  \textbs{99.218 $\pm$ 0.219}   & \textbs{68.555 $\pm$ 0.624}     & \textbs{0.853 $\pm$ 0.001}    &         0.187 $\pm$ 0.005     & \textgd{0.0901}                \\                                                                  
    \midrule                                                                                                                                                                                                                                                  
    EGT                      & \textbs{86.821 $\pm$ 0.020}    & \textbs{79.232 $\pm$ 0.348}     &  \textbs{98.173 $\pm$ 0.087}   & \textbs{68.702 $\pm$ 0.409}     & \textbs{0.853 $\pm$ 0.001}    & \textbs{0.108 $\pm$ 0.009}    & \textbs{0.0899}                \\
    \bottomrule   
\end{tabular}%
  }
  \label{tab:ablation}
\end{table*}
The results for the graph regression task on the OGB-LSC PCQM4M and PCQM4Mv2 datasets \cite{hu2021ogb} are presented in Table~\ref{tab:pcqm4m_res}. We show results for EGT models of small, medium and large network sizes based on number of parameters (details are included in the appendix). Note that the PCQM4M dataset was later deprecated in favor of PCQM4Mv2. So its test labels are no longer available and results are given over the validation set. We include these results for a thorough comparison with established models that report their results on the older dataset. We see that EGT achieves a much lower MAE than \emph{all} the convolutional and local self-attention based (i.e., GT \citeGT) GNNs. Its performance even exceeds Graphormer \citep{ying2021transformers}, which is also a global self-attention based model and can be thought of as an ablated variant of EGT with specialized encodings, such as centrality, spatial and edge encodings and requires similar training time and resources. We hypothesize that EGT gets a better result than Graphormer because of a combination of several factors, including its edge channels, unique gating mechanism and dynamic centrality scalers. Our model is currently the best performing model on the PCQM4Mv2 leaderboard. These results show the scalability of our framework and further confirm that given enough data, global self-attention based aggregation can outperform local convolutional aggregation.

\subsection{Transfer Learning Performance}
In order to experiment on the transferability of the representations learned by EGT, we take an EGT model pre-trained on the large-scale PCQM4Mv2 molecular dataset and fine-tune the weights on the OGB molecular datasets MolPCBA and MolHIV. Although the validation performance improvement seems to plateau for larger models on the PCQM4Mv2 dataset at a certain point, we found that larger pre-trained models perform better when fine-tuned on smaller datasets, so we select the largest model (EGT\textsubscript{Larger}) with 30 layers for transfer learning experiments (it achieves a validation MAE of 0.0869 on PCQM4Mv2, same as EGT\textsubscript{Large}). The results are presented in Table~\ref{tab:ogb_res}. We see that both EGT and Graphormer achieve comparable results which exceed convolutional GNNs. Graphormer uses pre-trained models from PCQM4M and they found it essential to use the FLAG training method \cite{kong2020flag} to achieve good fine-tuning results. FLAG uses an inner optimization loop to augment the node embeddings by adding adversarial perturbations to them. However, we do not use any form of specialized training during the fine-tuning process. This is due to two reasons - firstly, we wanted to evaluate our model in the conventional transfer learning setting where the weights of a pre-trained model are simply fine-tuned on a new dataset for a very few epochs which saves training time and resources -- whereas, FLAG training takes several times longer training time with additional FLAG hyperparameter tuning. Another reason is that FLAG is an adversarial perturbation method for node embeddings and since we have both node and edge embeddings (including non-existing edges) it is not clear how this method should be adopted for our model -- which requires further research.

\subsection{Ablation Study}
Our architecture is based upon two important ideas -- global self-attention based aggregation and residual edge channels. To analyze the importance of these two features, we experiment with two ablated variants of EGT: i) {\bf EGT-Simple:} incorporates global self-attention, but instead of having dedicated residual channels for edges, it directly uses a linear transformation of the input edge embeddings $e_{ij}^0$ (formed from adjacency matrix and edge features) to guide the self-attention mechanism. The absence of edge channels means that the edge embeddings $e_{ij}$ are not updated from layer to layer. So, edge classification is performed by applying MLP layers on pairwise node-embeddings. It is architecturally similar to Graphormer \citep{ying2021transformers}. While it is slightly less expensive in terms of computation and memory, it still scales quadratically with the number of nodes. ii) {\bf EGT-Constrained} limits the self-attention process to the 1-hop neighborhood of each node, which allows us to compare global self-attention to convolutional local self-attention based aggregation.  Also, it only keeps track of the edge embeddings $e_{ij}$ in the edge channels if there is an edge from node $i$ to node $j$ or $i=j$ (self-loop). Architecturally, this variant is similar to GT \citep{dwivedi2020generalization} and can take advantage of the sparsity of the graph to reduce computational and memory costs. More details about these variants can be found in the appendix.
\begin{table}[!t]
  \centering
  \caption{Ablation study on the PCQM4Mv2 dataset for EGT\textsubscript{Small} (from Table~\ref{tab:pcqm4m_res}).}
  \scalebox{0.77}{
    \begin{tabular}{ccccc|c}
    \toprule              
    \textbf{Gated}      & \textbf{Attention}  & \textbf{Virtual}    & \textbf{Centrality}  & \textbf{Positional}    & \textbf{Validate}         \\
    \textbf{Aggregation}& \textbf{Dropout}    & \textbf{Nodes}      & \textbf{Scalers}     & \textbf{Encodings}     & \textbf{MAE} $\downarrow$ \\
    \midrule\midrule                                 
    --                  & --                  & --                  &  --                  &    --                  & 0.0965                    \\
    \midrule
    \checkmark          & --                  & --                  &  --                  &    --                  & 0.0943                    \\
    \checkmark          & \checkmark          & --                  &  --                  &    --                  & 0.0926                    \\
    \checkmark          & \checkmark          & \checkmark          &  --                  &    --                  & 0.0919                    \\
    \checkmark          & \checkmark          & \checkmark          &  \checkmark          &    --                  & \textgd{0.0900}           \\
    \checkmark          & \checkmark          & \checkmark          &  \checkmark          &    \checkmark          & \textbs{0.0899}           \\
    \bottomrule
\end{tabular}
  }
  \label{tab:pcq_ablation}
\end{table}
\begin{figure*}[!t]
  \centering
  \includegraphics[width=\textwidth]{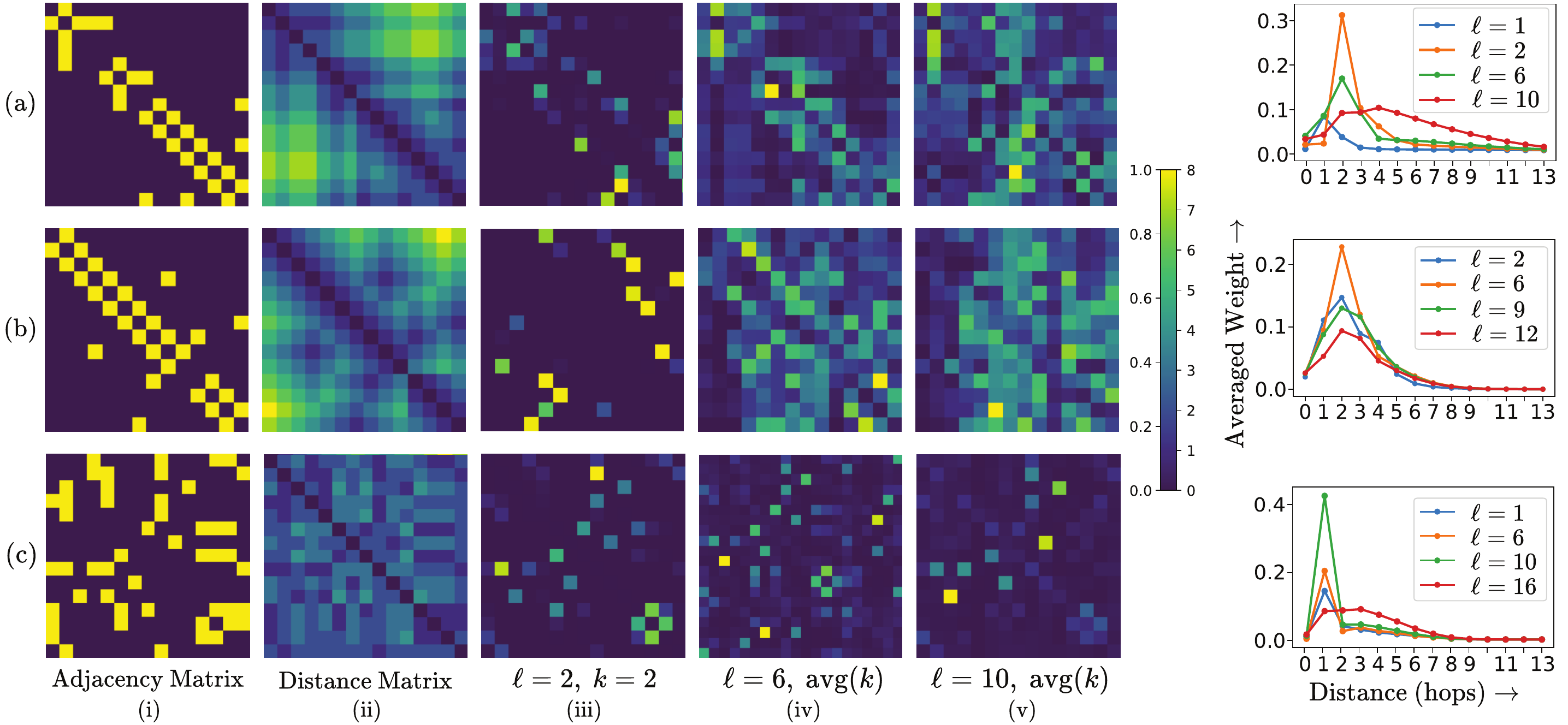}
  \caption{Analysis of aggregation patterns on three datasets -- (a) ZINC, (b) PCQM4Mv2, (c) TSP. Left to right -- adjacency (i) and distance matrices (ii), an example attention head (iii), average of attention heads in a middle layer (iv) and in a deeper layer (v) -- for a particular input graph in the validation set (matrices have been cropped for the TSP dataset). On the right -- weights assigned for different hops in different layers, averaged over all heads and all nodes in all the graphs in the validation set.}
  \Description{A visualization of attention patterns formed by EGT on three different datasets along with plots showing weights applied to different hops in different layers. These diagrams demonstrate how global self-attention based aggregation leads to learned non-local and sparse interaction between nodes in a graph.}
  \label{fig:att}
\end{figure*}

The results for the ablated variants are presented in Table~\ref{tab:ablation}. We see that, EGT-Simple can come close to EGT, but is especially subpar when the targeted task is related to edges (e.g., edge classification on the TSP dataset) or when the distance objective cannot be applied (ZINC, CIFAR10) due to the lack of dedicated edge channels. Both EGT-Simple and EGT enjoy an advantage over EGT-Constrained on the large PCQM4Mv2 dataset due to their global aggregation mechanism. This indicates that given enough data, global self-attention based aggregation can outperform local self-attention based aggregation. Additionally to demonstrate the effect of the SVD based positional encodings we include results without positional encodings. Note that the positional encodings lead to a significant improvement for the ZINC and the CLUSTER datasets, but slight/no improvement in other cases. This is consistent with our statement that the positional encodings are optional for our model on some tasks, but their inclusion can often lead to a performance improvement.

To further examine the contribution of different features of our model we carried out a series of experiments on the PCQM4Mv2 dataset for the smallest EGT network. The results are presented in Table~\ref{tab:pcq_ablation}. We see that the use of gates during aggregation leads to a significant improvement. Another important contributing factor is dropout on the attention matrix which encourages the network to take advantage of long-distance interactions. The dynamic centrality scalers also help by making the network more expressive. Virtual nodes and positional encodings lead to a more modest performance improvement.

\subsection{Analysis of Aggregation Patterns}
To understand how global self-attention based aggregation translates to performance gains we examined the attention matrices dynamically formed by the network. These matrices dictate the weighted aggregation of the nodes and thus show how each node is looking at other nodes. This is demonstrated in Fig.~\ref{fig:att}. We show the adjacency matrix and the distance matrix to demonstrate how far each node is looking. First, we look at an example attention matrix formed by an attention head. Next, for the sake of visualization, we merge the attention matrices for different heads together by averaging and normalizing them to values between $[0\ 1]$. We do this for two different layers at different depths of the model. Note that these patterns are specific to a particular input graph -- since the aggregation process is dynamic they would be different for different inputs. To make a complete analysis of each layer's attention we also plot the weights assigned at different distances averaged over all the attention heads for all the nodes and all the graphs in a dataset. Note that a convolutional aggregation of immediate neighbors would correspond to non-zero weights being assigned to only $0$/$1$ hop.

We see that the attention matrices for individual attention heads are quite sparse. So, the nodes are selective about where to look. For the ZINC dataset, from Fig.~\ref{fig:att} (a), at layer $\ell=1$ we see that EGT approximately follows a convolutional pattern. But as we go deeper, the nodes start to take advantage of global self-attention to look further. Finally, at $\ell=10$ we see highly non-local behavior. This shows why EGT has an advantage over local aggregation based convolutional networks because of its ability to aggregate global features. For PCQM4Mv2, in Fig.~\ref{fig:att} (b), we notice such non-local aggregation patterns starting from the lowest layers. This shows why a global aggregation based model such as EGT has a clear advantage over convolutional networks (as seen in Table~\ref{tab:pcqm4m_res}), because it would take a large number of consecutive convolutions to mimic such patterns. This non-local behavior is more subtle in TSP, where, except for the last layer, attention is mostly constrained to 1-3 hops, as seen from Fig.~\ref{fig:att}(c). This also shows why  EGT-Constrained achieves good results on this dataset (Table~\ref{tab:ablation}). However, even the slight advantage of global aggregation gives pure EGT an edge over EGT-constrained. To conclude, the aggregation performed by our model is sparse and selective, like convolution, yet capable of being non-local and dynamic, which leads to a clear advantage over convolutional networks.

\section{Conclusion and Future Work}
We proposed a simple extension -- edge channels -- to the transformer framework. We preserve the key idea, namely, global attention, while making it powerful enough to take structural information of the graph as input and also to process it and output new structural information such as new links and edge labels. One of our key findings is that the incorporation of the convolutional aggregation pattern is not an essential inductive bias for GNNs and instead the model can directly learn to make the best use of structural information. We established this claim by presenting experimental results on both medium-scale, large-scale and transfer learning settings where our model achieves superior performance, beating convolutional GNNs. We also achieve a new state-of-the-art result on the large-scale PCQM4Mv2 molecular dataset. We demonstrated that the performance improvement is directly linked to the non-local nature of aggregation of the model. In future work, we aim to evaluate the performance of EGT in transductive, semi-supervised and unsupervised settings. Also, we plan to explore the prospect of reducing the computation and memory cost of our model to a sub-quadratic scale by incorporating linear attention \citep{choromanski2020rethinking,katharopoulos2020transformers,schlag2021linear} and sparse edge channels.

\begin{acks}
This work was supported by the Rensselaer-IBM AI Research Collaboration, part of the IBM AI Horizons Network.
\end{acks}

\bibliographystyle{ACM-Reference-Format}
\bibliography{citations}

\appendix

\section{Data and Code Avalability}
\textbf{Data:} All datasets used in this work are publicly available. The medium-scale GNN benchmarking datasets by \citet{dwivedi2020benchmarking} are available at \url{https://github.com/graphdeeplearning/benchmarking-gnns}. The OGB-LSC \cite{hu2020open} PCQM4M and PCQM4Mv2 large-scale datasets, and the OGB datasets \citep{hu2021ogb} MolPCBA and MolHIV are available at \url{https://ogb.stanford.edu}.

\noindent\textbf{Code:} The code to reproduce the results presented in this work is available at \url{https://github.com/shamim-hussain/egt}.
\section{Training Method and Hyperparameters}
\begin{table}[!b]
	\caption{\small Common hyperparameters used in medium-scale experiments on all datasets.}
	\centering
	\scalebox{0.77}{
    \begin{tabular}{l|c}
    \toprule
    \textbf{Hyperparameter}                             & \textbf{Value}                                \\
    \midrule\midrule
    Number of attention heads, $H$                      & 8                                             \\
    Node channels FFN multiplier                        & 2                                             \\
    Edge channels FFN multiplier                        & 2                                             \\
    Final (two) MLP layers dimension                    & $d_h/2,d_h/4$                                 \\
    Virtual nodes                                       & Not used                                      \\
    SVD encoding rank, $r$                              & 8                                             \\
    Random attention masking rate                       & 0.1                                           \\
    Dynamic Centrality Scalers                          & Not used                                      \\
    Dropout                                             & Not used                                      \\
    Adam: initial LR                                    & $5\times10^{-4}$                              \\
    Adam: $\beta_1$							  	     	& 0.9                                           \\
    Adam: $\beta_2$                                     & 0.999                                         \\
    Adam: $\epsilon$                                    & $10^{-7}$                                     \\
    Reduce LR by factor                                 & 0.5                                           \\
    Minimum LR                                          & $5\times10^{-6}$                              \\
    LR warmup                                           & Not used                                      \\
    Cosine decay                                        & Not used                                      \\
    \bottomrule
\end{tabular}%
	}
	\label{tab:med_scale_hyper}
\end{table}

\begin{table*}[!t]
	\caption{\small Specific hyperparameters used for each dataset in medium-scale experiments. $\mathbf{D}^{(16)}$ is the distance matrix clipped to 16 hops. $\mathbf{A}$ is the adjacency matrix with self-loops. Distance prediction objective is only used for MNIST, CIFAR10 and ZINC datasets.}
	\centering
	\scalebox{0.77}{
	  \begin{tabular}{l|cc|c|c|c|cc|cc}
    \toprule
                                                        & \multicolumn{2}{c|}{\textbf{PATTERN}}                         & \textbf{CLUSTER}              & \textbf{MNIST}                & \textbf{CIFAR10}              & \multicolumn{2}{c|}{\textbf{TSP}}                             & \multicolumn{2}{c}{\textbf{ZINC}}               \\
    \cmidrule{2-10}
                                                        & \textbf{\#Param}              & \textbf{\#Param}              & \textbf{\#Param}              & \textbf{\#Param}              & \textbf{\#Param}              & \textbf{\#Param}              & \textbf{\#Param}              & \textbf{\#Param}       & \textbf{\#Param}       \\
    \textbf{Hyperparameter}                             & \textbf{$\approx$100K}        & \textbf{$\approx$500K}        & \textbf{$\approx$500K}        & \textbf{$\approx$100K}        & \textbf{$\approx$100K}        & \textbf{$\approx$100K}        & \textbf{$\approx$500K}        & \textbf{$\approx$100K} & \textbf{$\approx$500K} \\
    \midrule\midrule
    Input structural matrix                             & $\mathbf{D}^{(16)}$           & $\mathbf{D}^{(16)}$           & $\mathbf{D}^{(16)}$           & $\mathbf{A}$                  & $\mathbf{A}$                  & $\mathbf{D}^{(16)}$           & $\mathbf{D}^{(16)}$           & $\mathbf{A}$           & $\mathbf{A}$           \\
    Batch size                                          & 128                           & 128                           & 128                           & 128                           & 128                           & 8                             & 8                             & 128                    & 128                    \\
    Maximum no. of epochs                               & 200                           & 200                           & 200                           & 200                           & 200                           & 200                           & 200                           & 600                    & 600                    \\
    Reduce LR patience (epochs)                         & 10                            & 10                            & 10                            & 10                            & 10                            & 5                             & 5                             & 20                     & 20                     \\
    Distance prediction objective: $\nu$ (when used)    &                               &                               &                               & 3 hops                        & 3 hops                        &                               &                               & 3 hops                 & 3 hops                 \\
    Distance prediction objective: $\kappa$ (when used) &                               &                               &                               & $5\times10^{-4}$              & $5\times10^{-4}$              &                               &                               & $5\times10^{-2}$       & $5\times10^{-2}$       \\
    Number of layers, $L$                               & 4                             & 16                            & 16                            & 4                             & 4                             & 4                             & 16                            & 4                      & 10                     \\
    Node channels width, $d_h$                          & 64                            & 64                            & 64                            & 64                            & 48                            & 64                            & 64                            & 48                     & 64                     \\
    Edge channels width, $d_e$                          & 8                             & 8                             & 8                             & 8                             & 48                            & 8                             & 8                             & 48                     & 64                     \\
    \bottomrule
\end{tabular}%
	}
	\label{tab:med_scale_spec_hyper}
\end{table*}
\begin{table}[!t]
	\caption{\small Hyperparameters used in large-scale experiments.}
	\centering
	\scalebox{0.77}{
    \begin{tabular}{l|c}
    \toprule
    \textbf{Hyperparameter}                             & \textbf{Value}                                \\
    \midrule\midrule
    Input structural matrix                             & Distance matrix                               \\
                                                        & (clipped up to 16 hops)                       \\
    Number of attention heads, $H$                      & 32                                            \\
    Edge channels width, $d_e$                          & 64                                            \\
    Node channels FFN multiplier                        & 1                                             \\
    Edge channels FFN multiplier                        & 1                                             \\
    Final (two) MLP layers dimension                    & $d_h,d_h$                                     \\
    Virtual nodes                                       & 4                                             \\
    Dynamic Centrality Scalers                          & Used                                          \\
    SVD encoding rank, $r$                              & 8                                             \\
    Distance prediction objective                       & Not used                                      \\
    Random attention masking                            & Not used                                      \\
    Attention matrix dropout rate                       & 0.3                                           \\
    Adam: $\beta_1$							  	     	& 0.9                                           \\
    Adam: $\beta_2$                                     & 0.999                                         \\
    Adam: $\epsilon$                                    & $10^{-7}$                                     \\
    Reduce LR on loss plateau                           & Not used                                      \\
    Minimum LR                                          & $1\times10^{-6}$                              \\
    Batch size                                          & 512                                           \\
    LR warmup                                           & 200,000 steps                                 \\
    Cosine decay                                        & 800,000 steps                                 \\
    \midrule
    \textbf{Specific to EGT\textsubscript{Small}}       &                                               \\
    Maximum LR                                          & $2\times10^{-4}$                              \\
    Number of layers, $L$                               & 6                                             \\
    Node channels width, $d_h$                          & 512                                           \\
    \midrule
    \textbf{Specific to EGT\textsubscript{Medium}}      &                                               \\
    Maximum LR                                          & $1\times10^{-4}$                              \\
    Number of layers, $L$                               & 18                                            \\
    Node channels width, $d_h$                          & 640                                           \\
    \midrule
    \textbf{Specific to EGT\textsubscript{Large}}       &                                               \\
    Maximum LR                                          & $1\times10^{-4}$                              \\
    Number of layers, $L$                               & 24                                            \\
    Node channels width, $d_h$                          & 768                                           \\
    \midrule
    \textbf{Specific to EGT\textsubscript{Larger}}      &                                               \\
    Maximum LR                                          & $8\times10^{-5}$                              \\
    Number of layers, $L$                               & 30                                            \\
    Node channels width, $d_h$                          & 768                                           \\
    \bottomrule
\end{tabular}%
	}
	\label{tab:large_scale_hyper}
\end{table}
\begin{table}[!t]
	\caption{\small Hyperparameters used in transfer learning experiments.}
	\centering
	\scalebox{0.77}{
    \begin{tabular}{l|c|c}
    \toprule
    \textbf{Hyperparameter} & \textbf{MolPCBA}   & \textbf{MolHIV}    \\
    \midrule\midrule
    Maximum LR              & $1\times10^{-4}$   & $1\times10^{-4}$   \\
    Minimum LR              & $1\times10^{-6}$   & $1\times10^{-6}$   \\
    Batch size              & 16                 & 12                 \\
    LR warmup               & 20,000 steps       & 1,000 steps        \\
    Cosine decay            & 180,000 steps      & 2,000 steps        \\
    \bottomrule
\end{tabular}%
	}
	\label{tab:transfer_learn_hyper}
\end{table}
\subsection{Medium-scale Experiments}
For medium-scale experiments on the PATTERN, CLUSTER, MNIST, CIFAR10, TSP, and ZINC datasets we follow the benchmarking setting suggested by \citet{dwivedi2020benchmarking} and maintain a specified parameter budget of either 100K or 500K. The number of layers, the width of the node and the edge channels ($L$, $d_h$ and $d_e$, correspondingly) were varied to get the best results on the validation set. We used the Adam optimizer and reduce the learning rate by a factor of 0.5 if the validation loss does not improve for a given number of epochs (Reduce LR when validation loss plateaus). We keep track of the validation loss at the end of each epoch and pick the set of weights that produces the least validation loss. No dropout or weight decay is used for a fair comparison with other GNNs. Each experiment (training and evaluation) was run 4 times with 4 different random seeds and the results were used to calculate the mean and standard deviations of the metric. The common hyperparameters and methods for all datasets are given in Table~\ref{tab:med_scale_hyper}, whereas the hyperparameters which are specific for each dataset are given in Table~\ref{tab:med_scale_spec_hyper}.

\subsection{Large-scale Experiments}
While training large models on the PCQM4M and PCQM4Mv2 datasets, we found it essential to use learning rate warmup. Following the warmup, we applied cosine decay to the learning rate. We used virtual nodes which is a more scalable method than global average pooling because the use of multiple virtual nodes allows the model to collect more graph-level information. Instead of random masking of the attention matrices, we applied dropout to the attention matrices, which showed better regularization performance. Attention dropout is the only regularization method used for all models. We trained all models for a fixed number (1 million) of gradient update steps. The hyperparameters are shown in Table~\ref{tab:large_scale_hyper}.

\subsection{Transfer Learning Experiments}
We took the EGT\textsubscript{Larger} model pre-trained on the PCQM4Mv2 dataset (Table~\ref{tab:large_scale_hyper}) and fine-tuned it on the OGB datasets MolPCBA and MolHIV. We used the same learning rate and warmup and cosine decay method mentioned above, although for a smaller number of total gradient update steps. Hyperparameters specific to the fine-tuning stage are shown in Table~\ref{tab:transfer_learn_hyper}. Other hyper hyperparameters were the same as in Table~\ref{tab:large_scale_hyper}. Each experiment (training and evaluation) was run 4 times with 4 different random seeds and the results were used to calculate the mean and standard deviations of the metric.

\section{Details of Ablated Variants}
For the ablation study presented in section 4.4 of the paper, we discuss here different ablation methods in more detail.

\noindent\textbf{EGT-Simple:}
EGT-simple uses global self-attention, but does not have dedicated residual channels for updating pairwise information (edges). The input edge embeddings (formed from graph structural matrix and edge features) directly participate in the aggregation process as follows:
\begin{align}
  \tilde{\mathbf{A}}^{k,\ell} &= \mathrm{softmax} ( \hat{\mathbf{H}}^{k,\ell} ) \odot \sigma(\mathbf{G}^{k,\ell}_{0})  \\
  \text{Where,} \ \ \ \hat{\mathbf{H}}^{k,\ell}  &= \mathrm{clip}\left(\frac{\mathbf{Q}_h^{k,\ell} (\mathbf{K}_h^{k,\ell})^T}{\sqrt{d_k}}\right) + \mathbf{E}^{k,\ell}_{0}
\end{align}
$\mathbf{E}^{k,\ell}_{0}, \mathbf{G}^{k,\ell}_{0} \in \mathbb{R}^{N \times N}$ are directly formed by concatenations of the learned linear transformed input edge embeddings, i.e., $\mathbf{E}^{k,\ell} e_{ij}^{0}$, $\mathbf{G}^{k,\ell} e_{ij}^{0}$ respectively. Also, dynamic centrality scalers are derived from $e_{ij}^{0}$. The absence of edge channels means that the edge embeddings $e_{ij}$ are not updated from layer to layer. So, edge classification is performed from pairwise node embeddings and input edge features. We denote this variant as EGT-Simple since it is architecturally simpler than EGT.

We use the same hyperparameters for this variant as the ones used for original EGT (Table~\ref{tab:med_scale_spec_hyper}, Table~\ref{tab:large_scale_hyper}; $d_e$ denotes only the dimensionality of the input edge embeddings) except, $d_h=64$, $d_e=8$ for CIFAR10, and $d_h=80$, $d_e=8$ for ZINC are used to make the number of parameters comparable.

\noindent\textbf{EGT-Constrained:}
EGT-Constrained is a convolutional variant of EGT which limits the self-attention process to the 1-hop neighborhood of each node. It only keeps track of the edge embeddings $e_{ij}$ in the edge channels if there is an edge from node $i$ to node $j$ or $i=j$ (self-loop). So, pairwise information corresponding to only the existing edges is updated by the edge channels. This architecture can be derived by taking the softmax over $j\in \mathcal{N}(i)\cup \{i\}$ while calculating the attention weights $\tilde{\mathbf{A}}_{ij}^{k,\ell}$ and limiting the aggregation process to neighbors as:
\begin{align}
  \hat{\hat{h}}_i^{\ell} &= h_i^{\ell-1} + \mathbf{O}_h^\ell \bigparallel_{k=1}^{H} \sum_{j\in \mathcal{N}(i)\cup \{i\}}  \tilde{\mathbf{A}}_{ij}^{k,\ell} (\mathbf{V}^{k,\ell} \hat{h}_i^{\ell})
\end{align}
Since this architecture is constrained to the existing edges we denote this as EGT-Constrained. It has the advantage that depending on the sparsity of the graph, it can have sub-quadratic computational and memory costs. However, it can be difficult to perform sparse aggregation in parallel on the GPU. Instead of sparse operations, we used masked attention to implement this architecture for faster training on datasets containing smaller graphs because we can take advantage of highly parallel tensor operations.

For this variant, $d_h,d_e$ bear their usual meanings in the hyperparameters tables (Table~\ref{tab:med_scale_spec_hyper}, Table~\ref{tab:large_scale_hyper}). We use the same hyperparameters for this variant as the ones used for original EGT.

\noindent\textbf{Ungated Variant:}
In EGT, the edge channels participate in the aggregation process in two ways - by an attention bias and also by gating the values before they are aggregated by the attention mechanism. To verify the utility of the gating mechanism used in EGT, an ungated variant can be formulated by simplifying the aggregation process as follows: 
\begin{align}
  \tilde{\mathbf{A}}^{k,\ell} &= \mathrm{softmax} ( \hat{\mathbf{H}}^{k,\ell} )\\
  \text{Where,} \ \ \hat{\mathbf{H}}^{k,\ell}  &= \mathrm{clip}\left(\frac{\mathbf{Q}_h^{k,\ell} (\mathbf{K}_h^{k,\ell})^T}{\sqrt{d_k}}\right) + \mathbf{E}^{k,\ell}_{e}
\end{align}
$\mathbf{E}^{k,\ell}_{e} \in \mathbb{R}^{N \times N}$ is a concatenation of the learned linear transformed edge embeddings, i.e., $\mathbf{E}^{k,\ell} \hat{e}_{ij}^{\ell}$. Note that we omitted the sigmoid gates. The edge channels influence the aggregation process only via an attention bias.

\end{document}